\documentclass[11pt]{article}

\usepackage[preprint]{acl}

\usepackage{times}
\usepackage{latexsym}

\usepackage[T1]{fontenc}
\usepackage[utf8]{inputenc}

\usepackage{microtype}

\usepackage{inconsolata}

\usepackage{graphicx}

\author{
 \textbf{Jianqing Zhang\textsuperscript{1,2}},
 \textbf{Zhezheng Hao\textsuperscript{3}},
 \textbf{Wei Xia\textsuperscript{2}},
 \textbf{Hande Dong\textsuperscript{2}},
 \textbf{Hong Wang\textsuperscript{2}},
 \\
 \textbf{Chenxing Wei\textsuperscript{4}},
 \textbf{Yuyan Zhou\textsuperscript{2}},
 \textbf{Yubin Qi\textsuperscript{5}},
 \textbf{Qiang Lin\textsuperscript{2}},
 \textbf{Jian Cao\textsuperscript{1}},
\\
 \textsuperscript{1}Shanghai Jiao Tong University,
 \textsuperscript{2}Tencent, 
 \\
 \textsuperscript{3}Zhejiang University,
 \textsuperscript{4}Shenzhen University,
 \textsuperscript{5}Peking University
\\
 \small{
   \textbf{Correspondence:} 
   \href{mailto:tsingz@sjtu.edu.cn}{tsingz@sjtu.edu.cn},
   \href{mailto:xwell.xia@gmail.com}{xwell.xia@gmail.com},
   \href{mailto:cao-jian@sjtu.edu.cn}{cao-jian@sjtu.edu.cn}
 }
}

\usepackage{algorithm}
\usepackage{algorithmic}
\usepackage{bibentry}

\usepackage{xspace}
\def\gapo{\texttt{\textbf{GAPO}}\xspace}

\makeatletter
\DeclareRobustCommand\onedot{\futurelet\@let@token\@onedot}
\def\@onedot{\ifx\@let@token.\else.\null\fi\xspace}
\def\eg{\emph{e.g}\onedot} 
\def\ie{\emph{i.e}\onedot}

\makeatother

\definecolor{blue_}{RGB}{76, 114, 176}
\definecolor{orange_}{RGB}{221, 132, 82}
\definecolor{upload}{RGB}{47, 85, 151}
\definecolor{download}{RGB}{241, 13, 208}
\definecolor{red_}{RGB}{255, 0, 0}
\definecolor{gray_}{RGB}{127, 127, 127}
\definecolor{green_}{RGB}{1, 128, 0}
\definecolor{sjtured_}{RGB}{192, 0, 0}
\definecolor{sjtugreen_}{RGB}{84, 130, 53}
\definecolor{hist_red}{RGB}{194, 82, 83}
\definecolor{hist_blue}{RGB}{83, 110, 174}

\newcommand{\linecode}[1]{\colorbox[rgb]{1,1,1}{\color{black} \texttt{#1}}}

\usepackage{amsthm,amsmath,amssymb}
\usepackage{cleveref}

\crefname{section}{Sec.}{Secs.}
\crefname{table}{Table}{Tables}
\crefname{figure}{Figure}{Figures}
\crefname{equation}{Eq.}{Eqs.}

\usepackage{booktabs}
\usepackage{xcolor}

\usepackage{pifont}
\usepackage{makecell}
\usepackage{colortbl}
\definecolor{grayline}{gray}{0.9}

\usepackage{array}
\usepackage{multirow}
\usepackage{subfigure}
\usepackage[most]{tcolorbox}

\title{GAPO: Robust Advantage Estimation for Real-World Code LLMs}

\begin{document}

\maketitle

% \renewcommand{\thefootnote}{\fnsymbol{footnote}}
% \footnotetext[2]{Work done during the internship at Tencent.}
% \renewcommand{\thefootnote}{\arabic{footnote}}

\begin{abstract}
Reinforcement learning (RL) is widely used for post-training large language models (LLMs) in code editing, where group-relative methods, such as GRPO, are popular due to their critic-free and normalized advantage estimation. However, in real-world code-editing scenarios, reward distributions are often skewed with unpredictable noise, leading to distorted advantage computation and increased rollout outliers. To address this issue, we propose \textbf{Group Adaptive Policy Optimization (\gapo)}, which adaptively finds an \textit{interval with the highest SNR (signal-to-noise Ratio)} per prompt and uses the median of that interval as an adaptive $Q$ to replace the group mean in advantage calculation to reduce noise further. This adaptive $Q$ robustly handles rollout noise while remaining plug-and-play and efficient. We evaluate \gapo on nine instruction-tuned LLMs (3B–14B) using a collected large dataset of 51,844 real-world, history-aware code-editing tasks spanning 10 programming languages. \gapo yields up to 4.35 in-domain (ID) and 5.30 out-of-domain (OOD) exact-match improvements over GRPO and its variant DAPO, while achieving lower clipping ratios and higher GPU throughput. 
% Code: \url{https://github.com/TsingZ0/verl-GAPO}.
Code: \url{https://anonymous.4open.science/r/verl-GAPO-007F}.
\end{abstract}

\section{Introduction}

With the rapid advancement of large language models (LLMs), artificial intelligence (AI)–assisted coding has emerged as a prominent subfield and practical application \cite{chen2021evaluating, li2022competition}, demonstrating proven value in improving software engineering efficiency \cite{peng2023impact, yeticstiren2023evaluating}. Most code LLMs undergo a post-training stage, and reinforcement learning (RL) is a widely used method \cite{christiano2017deep, wang2024enhancing, hao2025rethinking}.

\begin{figure*}[h]
\centering
\includegraphics[width=\linewidth]{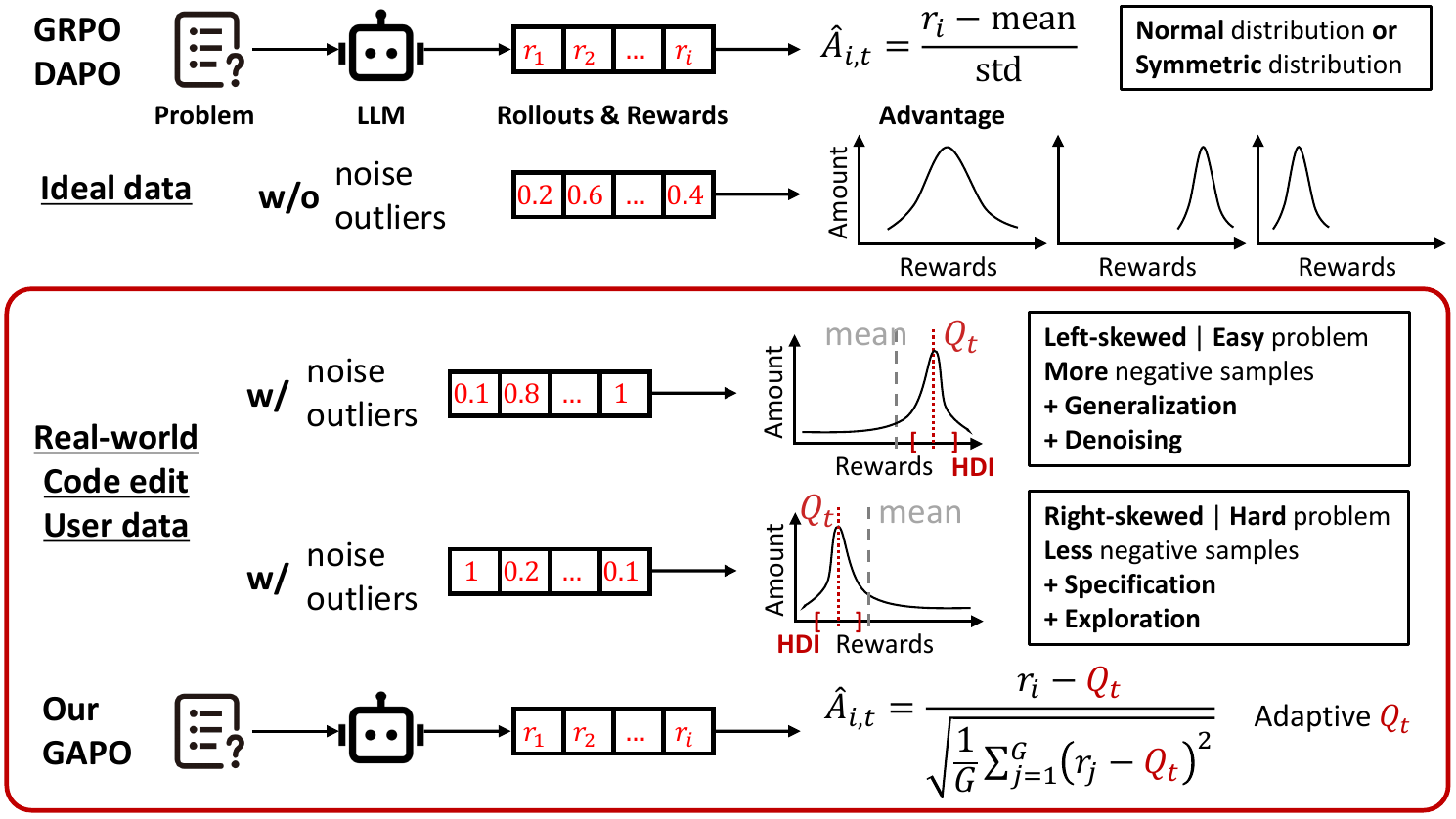}
\caption{Illustration of GRPO/DAPO and our proposed \gapo, where $r_i$ and $\hat{A}_{i,t}$ denote the $i$-th reward and the advantage within a group at step $t$, respectively. $Q_t$ is defined as the median of an adaptive \textit{Highest-Density Interval} (HDI) derived from the reward distribution of each prompt during the rollout process. }
\label{fig:intro}
\end{figure*}

Among RL methods, Group Relative Policy Optimization (GRPO) \cite{shao2024deepseekmath} and its variants are popular choices, known for their critic-free features \cite{liu2025part}. The core characteristic of the GRPO family lies in its group-relative advantage computation, which samples a group of rollouts for each input prompt and calculates advantage values as normalized rewards relative to the mean reward within each group \cite{shao2024deepseekmath}, as illustrated in \cref{fig:intro}. 

\begin{figure}[t]
\centering
\includegraphics[width=\linewidth]{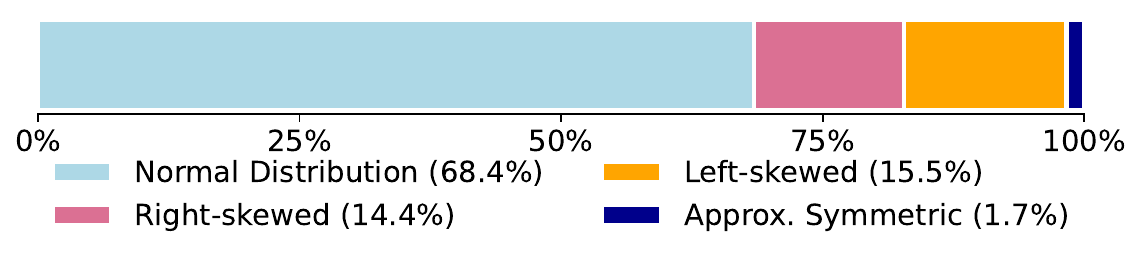}
\caption{Reward distribution of rollouts before training using Qwen2.5-Coder-14B \cite{hui2024qwen2}.}
\label{fig:bar}
\end{figure}

However, in real-world code-editing scenarios with complex contexts, inter-module invocation relationships, and diverse user intents, input prompts can inevitably induce noisy rollouts that contain outliers, which are unpredictable \cite{frauenknechtrollouts,wu2022plan}. 
In such practical cases, the expected normal or symmetric reward distributions often shift toward left- or right-skewed forms \cite{moore2009introduction}, a phenomenon frequently observed in practice (see \cref{fig:bar}). Given a normally distributed reward within the range $[0,1]$, when most rewards are greater than 0.5 (\ie, the mean exceeds 0.5), the practical reward distribution becomes left-skewed, as unpredictable outliers appear anywhere within the range, but their impact is more pronounced in the lower-reward region (below 0.5). The reverse occurs for right-skewed cases. Such rollout noise, often model- and scenario-specific, hinders LLM generalization and is hard to handle consistently. 

To reduce the impact of rollout noise and outliers, we propose \textbf{Group Adaptive Policy Optimization (\gapo)}, a robust group-adaptive advantage method that enhances GRPO and its variants (\eg, DAPO \cite{yu2025dapo}), as shown in \cref{fig:intro}. Unlike the mean, which treats all rewards uniformly and is sensitive to outliers, \gapo first identifies the sub-region with the highest signal-to-noise ratio (SNR) by reformulating the task as a classical \textit{Highest-Density Interval} (HDI) problem \cite{o2022smallest}, solved via an adapted sliding-window scan algorithm. We further enhance robustness by using the median of this region as the adaptive $Q$ instead of the mean in group-relative advantage computation. 

As shown in \cref{fig:intro}, the adaptive $Q$ provides additional benefits. For left-skewed distributions, \gapo generates more negative rollouts, improving generalization on easy problems \cite{mu2025dissecting, zhu2025surprising}. For right-skewed distributions, it promotes specialized learning on hard cases. The LLM-specific rollout noise also carries useful information, reflecting corner-case behaviors and forming the model’s \textit{blurry} ability edge. After updates, adaptive $Q$ suppresses noise in easy cases and amplifies outlier advantages in hard cases, enhancing the LLM’s ability edge. 

We evaluate nine diverse large language models (LLMs), both general-purpose and code-specialized, ranging from 3B to 14B parameters. 
Lacking public datasets for realistic, history-aware code edits, we collected 51,844 tasks across 10 languages—mainly Go (37.71\%), Python (22.14\%), and Java (21.03\%), each with a prompt (context, history, edit range) and the ground-truth edited snippet (see \cref{tab:data_attribute}). 
Extensive experiments demonstrate the superiority of our \gapo over both GRPO and DAPO in both accuracy and GPU throughput. In summary, our contributions are:
\begin{itemize}
    \item We are the first to observe that group-relative advantage is highly sensitive to outliers in real-world code editing, where reward distributions are often skewed by rollout noise.
    
    \item We introduce \gapo, which adaptively identifies the sub-region with the highest SNR for each input and scenario, enhancing the robustness of group-relative advantage calculation.
    
    \item We collect a large-scale real-world code-editing dataset and demonstrate \gapo’s ID and OOD superiority over GRPO and DAPO across nine LLMs (3B–14B), with lower clipping ratios and higher GPU throughput.
\end{itemize}

\section{Preliminaries}

\subsection{Problem Formulation}

In code editing, the model receives a prompt $p$ with context, history, edit region, cursor, user instructions, and other details. The LLM ($\theta$) generates a snippet $\hat{e}(p, \theta)$ to replace the edit region, or applies ``no change'' if indicated by special output tokens. 

During training, we have access to the ground-truth edit $e^*$, the code modification the user intended for each prompt $p$. 
The objective is to maximize the expected reward:
\begin{equation}
    \mathcal{J}(\theta) = \mathbb{E}_{p \sim P(\mathcal{P})} [\, r(\hat{e}(p, \theta), e^*) \,], \label{eq:problem}
\end{equation}
where $r(\cdot, \cdot)$ is a reward function that quantifies the similarity or correctness of the predicted edit relative to the ground-truth edit.

\subsection{Group Relative Policy Optimization}

A key advantage of GRPO is that it eliminates the need for a separate value function approximation; instead, it calculates advantages using the average reward across multiple sampled responses (rollouts) generated from the same input prompt. 
When using GRPO from an RL view, we can rewrite \cref{eq:problem} to be
\begin{equation}
\begin{aligned}
    \mathcal{J}_{GRPO}(\theta) = \mathbb{E}_{p \sim P(\mathcal{P}), \{e_i\}_{i=1}^G \sim \pi_{\theta_{old}}(O|p)}& \\
    \frac{1}{G} \sum_{i=1}^G \frac{1}{|e_i|} \sum_{i=1}^{|e_i|} [\min(\kappa_{i,t} \cdot \hat{A}_{i,t}, \rho_{i,t, \epsilon} \cdot \hat{A}_{i,t} &) + C], \\
    s.t., \quad \kappa_{i,t} = \frac{\pi_{\theta}(e_{i,t}|p, e_{i, <t})}{\pi_{\theta_{old}}(e_{i,t}|p, e_{i, <t})},& \\
    \rho_{i,t, \epsilon} = \text{clip}\Big(\frac{\pi_{\theta}(e_{i,t}|p, e_{i, <t})}{\pi_{\theta_{old}}(e_{i,t}|p, e_{i, <t})}, 1-\epsilon,& 1+\epsilon\Big), \\
    C = - \beta \mathbb{D}_{KL}[\pi_\theta || \pi_{ref}],
\end{aligned}
\end{equation}
where $\pi_{\theta}$, $\pi_{\theta_{old}}$, and $\pi_{ref}$ are the updating policy model, old policy model, and reference policy model, respectively. For each prompt $p$, GRPO samples a group of $G$ edits $\{e_1, e_2, \ldots, e_G\}$ from the old policy $\pi_{\theta_{old}}$. 
Here, $\epsilon$ and $\beta$ are preset hyperparameters that control the optimization behavior, and $\mathbb{D}_{\mathrm{KL}}$ denotes the KL divergence between the current training policy and a fixed reference policy \cite{shao2024deepseekmath}. 

The advantage $\hat{A}_{i,t}$ is computed based on the relative rewards of the edits within each group. Given any reward model, we score each edit to obtain a set of $G$ rewards $\textbf{r} = \{r_1, r_2, \ldots, r_G\}$, which are then normalized to yield the advantage calculation:
\begin{equation}
    \hat{A}_{i,t} = \frac{r_i - \text{mean}(\textbf{r})}{\text{std}(\textbf{r})}. \label{eq:adv_grpo}
\end{equation}

\section{Method}

\subsection{Motivation}

As shown in \cref{fig:bar}, many real-world code-editing prompts produce skewed reward distributions with outliers, adding noise and negative impact to RL rollouts and training \cite{frauenknechtrollouts, hollenstein2022action} because \cref{eq:adv_grpo} aggregates all rewards uniformly. 
Common methods use non-adaptive geometric-mean \cite{zhao2025geometric} or quantile statistics to reduce outlier influence \cite{john2015robustness, rousseeuw2011robust}, but as \cref{fig:bar} shows, reward distributions vary across prompts, so a single mean or quantile cannot adapt effectively.
Moreover, \textit{noise can be a form of LLM-specific useful information} that should not be discarded, as it reflects the model’s output and corner-case behaviors, forming its ‘blurry ability edge.’ 
To address these, we propose a group-adaptive advantage calculation that identifies the highest-density reward region, minimizing the impact of outliers, while also leveraging noise by amplifying the absolute advantage of outliers to make the LLM’s ability edge more pronounced.

\subsection{Group Adaptive Policy Optimization} 

Our \gapo method does not alter the objective of existing group-relative RL approaches; instead, it only modifies the advantage computation in \cref{eq:adv_grpo}. This design makes it simple to implement and plug-and-play with various group-relative RL frameworks, such as \texttt{verl} \cite{sheng2025hybridflow}, requiring only a few lines of code. Specifically, we redefine the advantage as
\begin{equation}
    \hat{A}_{i,t} = \frac{r_i - Q_t}{\sqrt{\frac{1}{G}\sum_{j=1}^G (r_j - Q_t)^2}}, \label{eq:adv_our}
\end{equation}
where the denominator represents a variant of the standard deviation, ensuring consistency with the replacement of the mean by the adaptive $Q$ value. 

The key challenge, then, is to obtain the adaptive $Q$ value for each group corresponding to each input prompt while updating the policy models. Although the median can mitigate the effect of outliers, it is computed over the entire group and is less adaptive to varying user scenarios. Thus, we propose first to \textit{identify the region free of outliers} and then \textit{set $Q$ as the median of this sub-region $\mathcal{H}$}. 

To achieve this, we seek a sub-region with the highest SNR, which represents the majority of the rewards with fewer outliers. This naturally leads to a classical statistical problem of finding the HDI for a given probability mass, that is, the narrowest interval \cite{o2022smallest}. Formally, for a group of reward values, this can be expressed as

\begin{tcolorbox}[colback=white, colframe=gray, coltitle=white, colbacktitle=black, fonttitle=\bfseries, title=Highest-Density Interval (HDI), rounded corners=southwest, boxrule=0.5mm, arc=2mm, width=\linewidth, breakable]
\small
    Given $G$ real numbers (rewards) $r_1, r_2, \ldots, r_G$ and an integer $k$ (here $k=\lceil G\tau \rceil$), find indices $i<j$ with $j-i+1 \ge k$ that minimize the interval length 
    $$
    L(i,j) = r_j - r_i,
    $$
    where $r_{(1)}, r_{(2)}, \ldots, r_{(G)}$ are the sorted values. 
\end{tcolorbox}

\begin{algorithm}[h]
\caption{Find the Highest-Density Interval}
\label{alg:shortest_segment}
\begin{algorithmic}[1]
\REQUIRE $\mathcal{G}$: $G$ reward values in $[0,1]$, $\tau \in [0,1]$. 
\ENSURE $\mathcal{H}$: the HDI covering at least $\tau$ of points.
\STATE $\mathcal{G}' \gets$ sort($\mathcal{G}$)
\STATE $min\_length \gets \infty$
\STATE $best\_start \gets 0$
\STATE $best\_end \gets k - 1$
\FOR{$start = 0$ \textbf{to} $G - k$}
    \STATE $end \gets start + k - 1$
    \STATE $current\_length \gets \mathcal{G}'[end] - \mathcal{G}'[start]$
    \IF{$current\_length < min\_length$}
        \STATE $min\_length \gets current\_length$
        \STATE $best\_start \gets start$
        \STATE $best\_end \gets end$
    \ENDIF
\ENDFOR
\STATE $\mathcal{H} \gets \mathcal{G}'[best\_start : best\_end+1]$
\RETURN $\mathcal{H}$
\end{algorithmic}
\end{algorithm}

Next, to solve this problem, we use the algorithm in \Cref{alg:shortest_segment}. The method first sorts the data to obtain order statistics, then applies a sliding window of fixed size $k$ over the sorted array, computing $r_{(i+k-1)} - r_{(i)}$ for each window. The minimal such difference corresponds to the shortest interval containing $k$ points.  

\textit{Why this is optimal}: any interval covering $k$ points corresponds to a contiguous block in the sorted list; enlarging the interval beyond $k$ points cannot make it strictly shorter, so it suffices to check only blocks of size exactly $k$. The computational complexity is dominated by the sorting step, which requires $\mathcal{O}(n \log n)$ time ($n$: number of rollouts, usually small), while the sliding window scan costs only $\mathcal{O}(n)$. Therefore, the total complexity is $\mathcal{O}(n \log n)$ time with $\mathcal{O}(m)$ extra space.

\section{Experiment}

\textbf{LLMs.} 
For comprehensive evaluation, we consider \textbf{\textit{nine}} large language models (LLMs) covering a diverse spectrum: \textbf{Mistral-v0.3} \cite{jiang2023mistral7b} (general-purpose, non-reasoning; 7B), \textbf{Qwen2.5} \cite{team2024qwen2} (general-purpose, non-reasoning; 3B, 7B), \textbf{Qwen3} \cite{yang2025qwen3} (general-purpose, reasoning; 4B, 8B), \textbf{Qwen2.5-Coder} \cite{hui2024qwen2} (code-specialized, non-reasoning; 3B, 7B, 14B), and \textbf{DeepSeek-Coder} \cite{guo2024deepseek} (code-specialized, non-reasoning; 6.7B). All these models are instruction-tuned versions downloaded from Hugging Face \cite{huggingface}. 

\textbf{Training and Evaluation Data.} 
Our goal is to work on code editing tasks in \textit{real-world software engineering} with rich context and editing histories, but no open-sourced dataset is available. Therefore, we collect training data from internal company users and remove all data containing private and sensitive information. Specifically, we finally have a total of 51,844 code-editing data points across 10 programming languages (Go, Python, Java, C++, Kotlin, TypeScript, JavaScript, C, Rust, and Lua). As shown in \cref{tab:lang}, Go, Python, and Java are the majority, which are also among the most popular programming languages \cite{tiobe_index}. 
Moreover, \cref{tab:lang} shows a large variation in input prompt and output editing lengths across scenarios, \textit{\textbf{requiring adaptation}}. For evaluation, we use the only available open-sourced zeta dataset \cite{zedindustries_zeta} (OOD), which contains hundreds of samples, together with our collected test set (ID) of 3,897 cases. 

\begin{table}[h] 
\caption{Statistics of the training data for real-world code editing tasks. “Input” and “Output” denote the input prompt and output ground-truth \textit{text length} ranges, respectively.
} 
\centering 
\resizebox{\linewidth}{!}{ 
\begin{tabular}{l|cc|cc} 
\toprule
 & \textbf{Input} & \textbf{Output} & \textbf{Count} & \textbf{Percent} \\
\midrule
Go          & 1,925 -- 24,883 & 36 -- 717 & 19,549 & 37.71\% \\
Python      & 1,984 -- 24,651 & 36 -- 788 & 11,477 & 22.14\% \\
Java        & 2,021 -- 23,181 & 38 -- 833 & 10,905 & 21.03\% \\
C++         & 2,008 -- 22,463 & 39 -- 736 & 3,245  & 6.26\%  \\
Kotlin      & 2,146 -- 16,353 & 38 -- 706 & 2,302  & 4.44\%  \\
TypeScript  & 2,033 -- 21,603 & 41 -- 557 & 1,946  & 3.75\%  \\
JavaScript  & 2,033 -- 18,230 & 43 -- 568 & 1,239  & 2.39\%  \\
C           & 2,134 -- 14,106 & 44 -- 628 & 896    & 1.73\%  \\
Rust        & 2,608 -- 11,204 & 44 -- 507 & 221    & 0.43\%  \\
Lua         & 2,991 -- 10,348 & 44 -- 611 & 64     & 0.12\%  \\
\midrule
Total & 1,925 -- 24,883 & 36 -- 833 & 51,844 & 100.00\% \\
\bottomrule
\end{tabular}} 
\label{tab:lang} 
\end{table}

\begin{table*}[h]
\caption{Exact match accuracy on the test set for nine LLMs. The asterisk (*) denotes reasoning LLMs. Step $\Delta$: \gapo requires fewer training steps than the baselines to reach their best accuracy, showing better training efficiency.}
\centering
\resizebox{\linewidth}{!}{
\begin{tabular}{l|c|cc|cc|ccc|c}
\toprule
Name & Mistral-v0.3 & \multicolumn{2}{c|}{Qwen2.5} & \multicolumn{2}{c|}{Qwen3*} & \multicolumn{3}{c|}{Qwen2.5-Coder} & DeepSeek-Coder \\
Size & 7B & 3B & 7B & 4B & 8B & 3B & 7B & 14B & 6.7B \\
\midrule
GMPO & 12.71 & 38.94 & 39.38 & 39.20 & 36.95 & 39.74 & 40.12 & 42.58 & 23.07\\
KRPO & 13.02 & 38.73 & 39.14 & 39.53 & 37.35 & 39.80 & 40.31 & 42.58 & 23.07\\
QAE & 16.52 & 32.99 & 39.47 & 38.25 & 39.76 & 38.34 & 41.89 & --- & 40.58\\
\midrule
GRPO & 12.93 & 38.80 & 39.05 & 39.47 & 36.80 & 39.69 & 40.05 & 42.64 & 23.32\\
\gapo (G) & 13.58 & 39.96 & 41.36 & 40.09 & 39.62 & 42.70 & 44.40 & 46.25 & 23.85\\
Improve. & \textcolor{green_}{\textbf{+0.65}} & \textcolor{green_}{\textbf{+1.16}} & \textcolor{green_}{\textbf{+2.31}} & \textcolor{green_}{\textbf{+0.62}} & \textcolor{green_}{\textbf{+2.82}} & \textcolor{green_}{\textbf{+3.01}} & \textcolor{green_}{\textbf{+4.35}} & \textcolor{green_}{\textbf{+3.61}} & \textcolor{green_}{\textbf{+0.53}}\\
Step $\Delta$ & -87 & -109 & -100 & -55 & -42 & -139 & -169 & -121 & -58\\
\midrule
DAPO & 16.59 & 32.80 & 39.46 & 37.99 & 39.80 & 38.74 & 41.64 & --- & 41.09\\
\gapo (D) & 17.20 & 33.87 & 41.13 & 39.62 & 40.64 & 39.98 & 43.96 & --- & 43.03\\
Improve. & \textcolor{green_}{\textbf{+0.61}} & \textcolor{green_}{\textbf{+1.07}} & \textcolor{green_}{\textbf{+1.67}} & \textcolor{green_}{\textbf{+1.37}} & \textcolor{green_}{\textbf{+0.84}} & \textcolor{green_}{\textbf{+1.24}} & \textcolor{green_}{\textbf{+2.32}} & --- & \textcolor{green_}{\textbf{+1.94}}\\
Step $\Delta$ & -58 & -80 & -33 & -19 & -48 & -93 & -65 & --- & -45 \\
\bottomrule
\end{tabular}}
\label{tab:nine}
\end{table*}

\textbf{Input-Output Structure.}
Each data consists of two fields: \linecode{<prompt>} and \linecode{<edit>}. The \linecode{<prompt>} field contains the code context, a sequence of edit histories, the code edit range (with the cursor's position), and user-provided hints. The \linecode{<edit>} field is the ground truth output. Details are summarized in \cref{tab:data_attribute}.

\begin{table}[h]
\centering
\resizebox{\linewidth}{!}{
\begin{tabular}{l|l}
\toprule
\textbf{Field} & \textbf{Components} \\
\midrule
\linecode{<prompt>} & \linecode{<system prompt>} \\
 & \linecode{<current code>} \\
 & \linecode{<sequence of edit histories>} \\
 & \linecode{<code edit range>} \& \linecode{<cursor>} \\
 & \linecode{<user prompt>} \\
\midrule
\linecode{<edit>} & \linecode{<ground truth>} \\
\bottomrule
\end{tabular}}
\caption{Components of each data in the training and test datasets.}
\label{tab:data_attribute}
\end{table}

\textbf{Baselines.} 
(1) Group Relative Policy Optimization (GRPO) \cite{shao2024deepseekmath} is a widely influential RL algorithm for LLMs \cite{kumar2025llm}. (2) Decoupled Clip and Dynamic Sampling Policy Optimization (DAPO) \cite{yu2025dapo} is a popular successor to GRPO. (3) GMPO \cite{zhao2025geometric} replaces the arithmetic mean with a non-adaptive geometric mean. (4) KRPO \cite{wang2025kalman} employs a Kalman filter to estimate latent reward uncertainty. (5) QAE \cite{wu2025quantile} introduces a group-wise $K$-quantile reward baseline for entropy-safe reasoning based on DAPO. 
GMPO, KRPO, and QAE are designed for discrete-reward tasks and \textit{lack adaptability to diverse real-world noisy rollout distributions}. 

\textbf{Other Settings.} 
We use the popular \texttt{verl} framework \cite{sheng2025hybridflow} for RL training, which is widely adopted in industry due to its scalability. We follow most of the default settings for GRPO and DAPO in \texttt{verl} with adaptive modifications for our code edit tasks.
Our \gapo method is straightforward to implement, requiring only a few lines of code by modifying the \linecode{compute\_grpo\_outcome\_advantage} function in the GRPO and DAPO implementations (see details in our code), showing its high compatibility. We use the exact match ($em$) as the evaluation metric following \cite{deng2025enhancing}. All results are reported as the average over five trials. See more details and results in the Appendix. 

We use a continuous-reward function that combines the exact match ($em$) metric \cite{dibia2023aligning} and a normalized edit distance ($ed$) metric:
\begin{equation}
    r(\hat{e}, e^*) = \frac{1}{2} \Big[em(\hat{e}, e^*) + (1 - \frac{ed(\hat{e}, e^*)}{\max\{l(\hat{e}), l(e^*)\}}) \Big], \label{eq:reward}
\end{equation}
where $\hat{e}$ and $e^*$ represent LLM output and ground truth, respectively. $ed(\hat{e}, e^*)$ is computed using the classical dynamic programming algorithm \cite{lcvenshtcin1966binary}, and $l(\cdot)$ returns the code length.

\subsection{In-Domain Performance}

We begin by presenting the improvements of our \gapo over both GRPO and DAPO on \textit{\textbf{nine}} LLMs, covering general-purpose and code-specific models, as well as reasoning and non-reasoning types. The original DAPO implementation encounters out-of-memory (OOM) issues with 14B models in our long-context setting, leading to missing results. \gapo (G) and \gapo (D) denote the application of our adaptive $Q$ to GRPO and DAPO, respectively. 

As shown in \cref{tab:nine}, our \gapo yields greater benefits for the Qwen series compared to Mistral and DeepSeek. Within the Qwen family, \gapo demonstrates the strongest effect on Qwen2.5-Coder, achieving up to a 4.35-point improvement in exact match accuracy over GRPO/DAPO. This indicates that \gapo is particularly effective for initially strong, code-specific LLMs. In contrast, its effectiveness is limited for weaker models with low baseline performance, such as Mistral-v0.3.
Overall, the \textit{\textbf{adaptive advantage computation}} at the core of \gapo generalizes seamlessly to both GRPO and DAPO. However, GMPO, KRPO, and QAE perform similarly to their baselines (GRPO or DAPO), showing unstable improvements on real-world tasks with diverse noisy reward distributions.

Beyond accuracy, \gapo achieves higher training efficiency (Step $\Delta$ in \cref{tab:nine}), requiring fewer steps than GRPO and DAPO to reach the same accuracy, with especially large gains over GRPO. This stems from improved generalization on easy problems and enhanced specialization on hard ones, aligned with the optimization objective. Using adaptive $Q$ encourages exploration on hard cases by amplifying the absolute advantage of outliers (with infrequent high rewards), making the gains more pronounced for GRPO, which lacks inherent exploration \cite{yu2025dapo}.

\subsection{Out-of-Domain Performance}

\begin{table}[h] 
\caption{Exact match accuracy on the OOD zeta set for Qwen2.5-Coder and Qwen3. } 
\centering 
\resizebox{\linewidth}{!}{ 
\begin{tabular}{l|cc|ccc} 
\toprule 
Model & \multicolumn{2}{c|}{Qwen3*} & \multicolumn{3}{c}{Qwen2.5-Coder} \\ 
Size & 4B & 8B & 3B & 7B & 14B \\ 
\midrule 
GMPO & 7.58 & 14.39 & 17.42 & 21.21 & 21.97 \\ 
KRPO & 8.33 & 15.15 & 18.18 & 22.73 & 23.49 \\ 
QAE & 13.64 & 18.94 & 14.39 & 16.67 & --- \\ 
\midrule 
GRPO & 8.33 & 13.64 & 17.42 & 21.97 & 22.73 \\ 
\gapo (G) & 10.61 & 16.67 & 20.45 & 24.24 & 25.76 \\ 
Improve. & \textcolor{green_}{\textbf{+2.27}} & \textcolor{green_}{\textbf{+3.03}}  & \textcolor{green_}{\textbf{+3.03}}  & \textcolor{green_}{\textbf{+2.27}}  & \textcolor{green_}{\textbf{+3.03}} \\ 
Imp. (\%) & \textcolor{green_}{\textbf{+27.27}} & \textcolor{green_}{\textbf{+22.22}}  & \textcolor{green_}{\textbf{+17.39}}  & \textcolor{green_}{\textbf{+10.34}}  & \textcolor{green_}{\textbf{+13.33}} \\ 
\midrule 
DAPO & 12.88 & 19.70 & 13.64 & 16.67 & --- \\ 
\gapo (D) & 16.67 & 22.73 & 18.94 & 20.45 & --- \\ 
Improve.& \textcolor{green_}{\textbf{+3.79}} & \textcolor{green_}{\textbf{+3.03}}  & \textcolor{green_}{\textbf{+5.30}}  & \textcolor{green_}{\textbf{+3.79}}  & --- \\ 
Imp. (\%) & \textcolor{green_}{\textbf{+29.41}} & \textcolor{green_}{\textbf{+15.38}}  & \textcolor{green_}{\textbf{+38.89}}  & \textcolor{green_}{\textbf{+22.73}}  & --- \\ 
\bottomrule 
\end{tabular}} 
\label{tab:zeta} 
\end{table}

We further evaluate \gapo on the open-sourced zeta dataset \cite{zedindustries_zeta}, which, to our knowledge, is the only code-editing benchmark currently available and is out-of-domain (OOD) relative to our collected training set. Because the zeta set is small, the accuracy exhibits a stepwise pattern. As shown in \cref{tab:zeta}, \gapo consistently outperforms the original GRPO/DAPO across model scales on this OOD benchmark, achieving gains of up to 5.30 absolute points and 38.88\% relative improvement. These substantial OOD improvements stem from \gapo’s robust feature and enhanced generalization capability. 

\subsection{Performance Curves}

% \begin{figure}[h]
% \centering
% \includegraphics[width=\linewidth]{figs/W_B_grpo.pdf}
% \caption{Performance curves on the test set for GRPO, as logged by W\&B. The \textit{dotted} curves correspond to LLMs post-trained with the original GRPO. The curves labeled with “median-div” correspond to our proposed \gapo.}
% \label{fig:curve_per_grpo}
% \end{figure}

\begin{figure}[h]
\centering
\includegraphics[width=\linewidth]{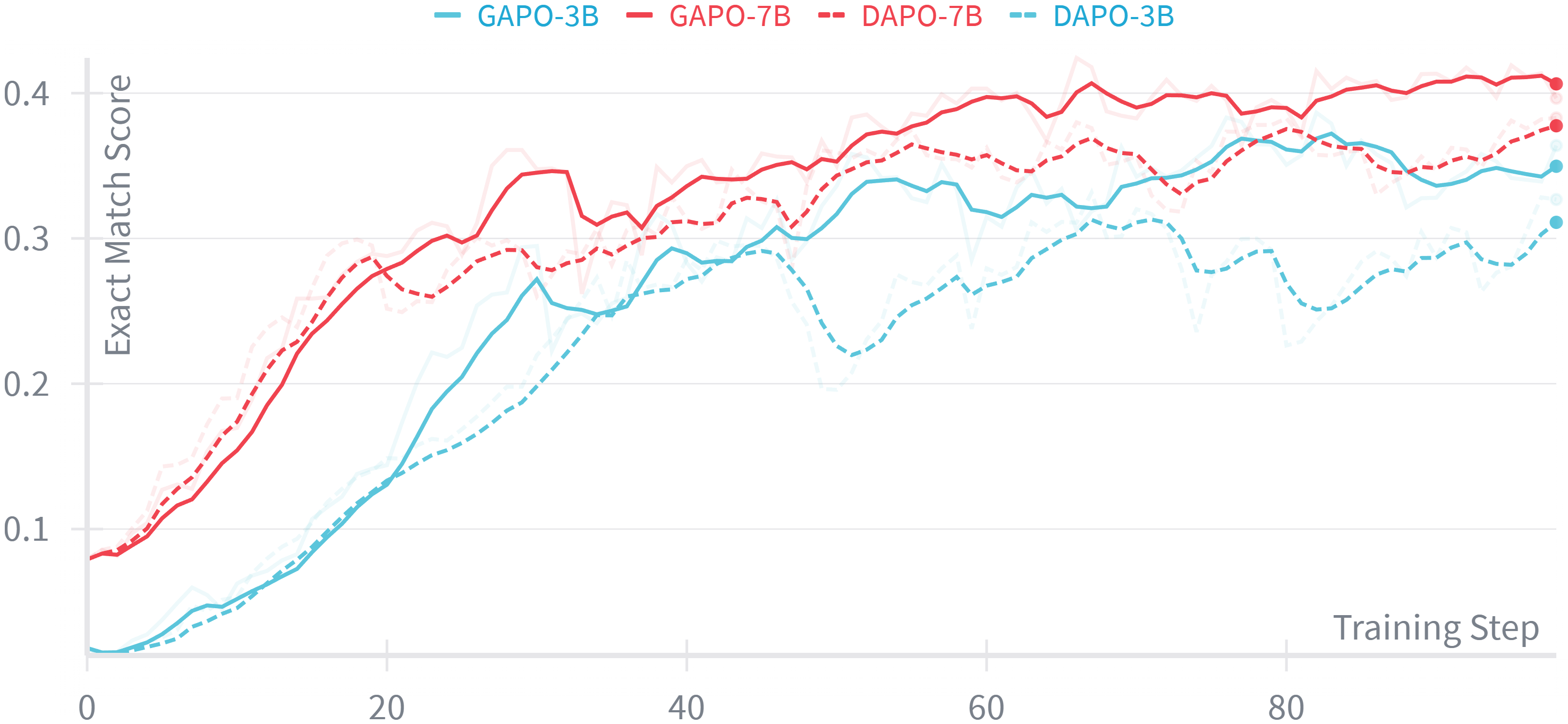}
\caption{Performance curves on the test set for DAPO with Qwen2.5-Coder-3B/7B, as logged by W\&B. The \textit{dotted} curves correspond to LLMs post-trained with the original DAPO. }
\label{fig:curve_per_dapo}
\end{figure}

We further present the W\&B-logged ID performance curves in \cref{fig:curve_per_dapo}. Due to space limitations, only the results for DAPO are shown. The curves demonstrate that \gapo consistently improves DAPO, with clear performance gains that persist during training. Notably, \gapo complements DAPO’s dynamic sampling, which filters zero-variance prompts and increases the relative proportion of outlier prompts.

\subsection{Policy Training Curves}

% \begin{figure}[h]
% \centering
% \includegraphics[width=\linewidth]{figs/W_B_grpo-actor.pdf}
% \caption{Clip fraction curves on the training set for GRPO, as logged by W\&B. }
% \label{fig:curve_clip_grpo}
% \end{figure}

\begin{figure}[h]
\centering
\includegraphics[width=\linewidth]{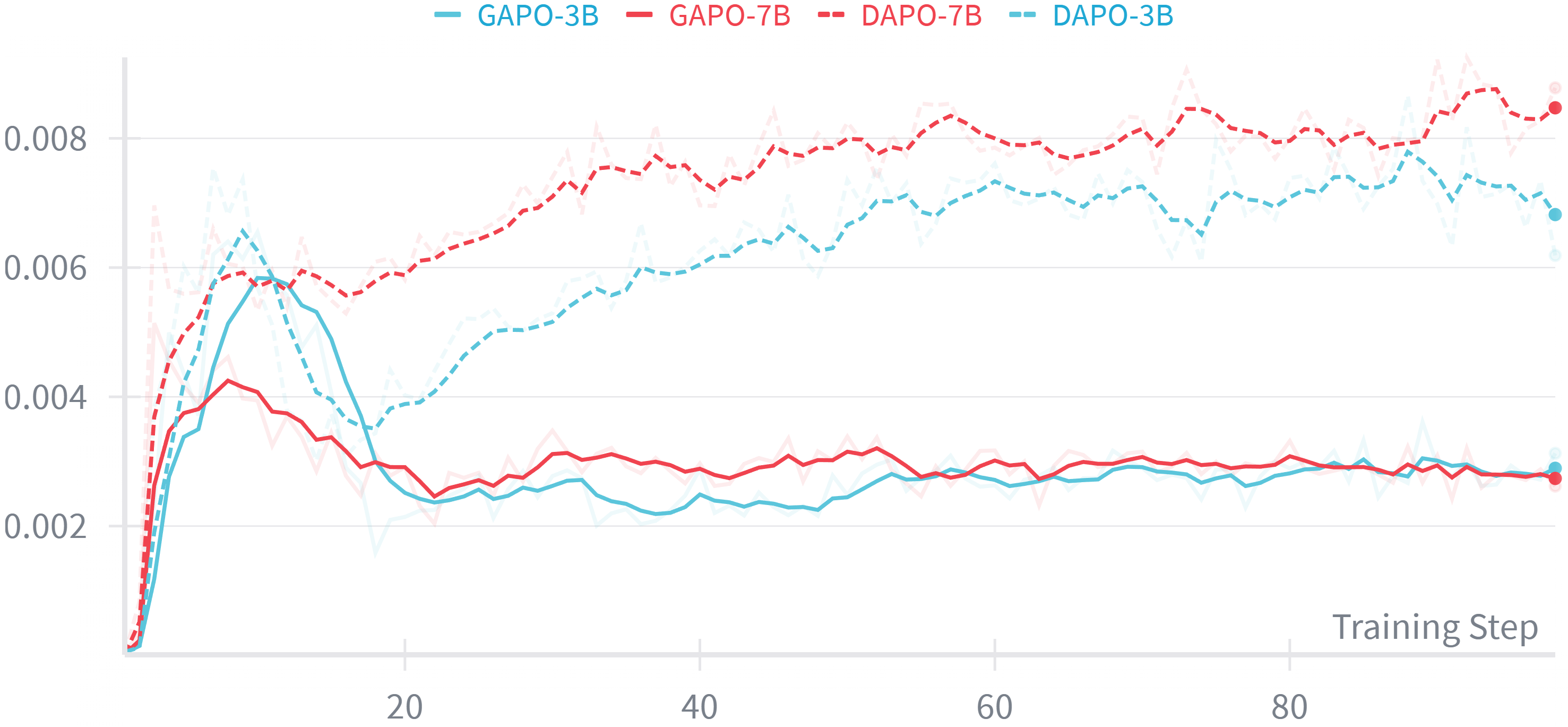}
\caption{Clip fraction curves on the training set for DAPO with Qwen2.5-Coder-3B/7B, logged by W\&B. }
\label{fig:curve_clip_dapo}
\end{figure}

The clip fraction curves in \cref{fig:curve_clip_dapo} illustrate the proportion of actions whose probability ratios were clipped during gradient updates. A low \texttt{pg\_clipfrac} indicates that few updates reach the clipping threshold, implying gentler policy changes and better alignment with the model’s current capabilities, allowing fuller utilization of the reward information. For both the peak and converged values, our \gapo consistently exhibits lower \texttt{pg\_clipfrac} than DAPO, especially in later training steps.

\subsection{GPU Throughput Curves}

% \begin{figure}[h]
% \centering
% \includegraphics[width=\linewidth]{figs/W_B_grpo-throughput.pdf}
% \caption{GPU throughput curves on the training set for GRPO, as logged by W\&B. }
% \label{fig:throughput_grpo}
% \end{figure}

\begin{figure}[h]
\centering
\includegraphics[width=\linewidth]{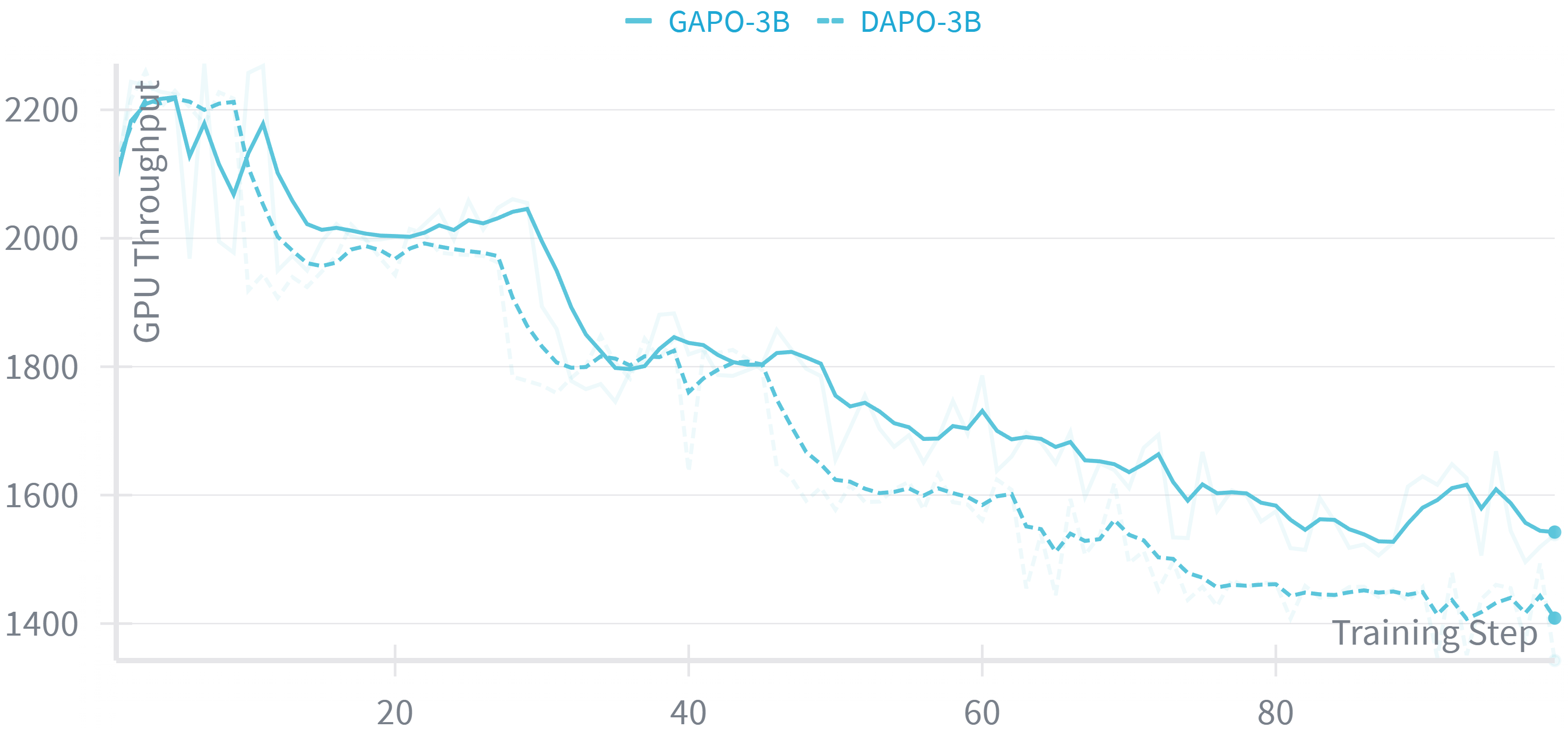}
\caption{GPU throughput curves on the training set for DAPO with Qwen2.5-Coder-3B, as logged by W\&B. }
\label{fig:throughput_dapo}
\end{figure}

GPU throughput is closely related to training efficiency, typically measured as the number of tasks processed per unit time. Due to magnitude differences, we present results only for the 3B models. By reducing noise and enhancing training stability, \gapo indirectly improves GPU throughput, as shown in \cref{fig:throughput_dapo}. Quantitatively, \gapo improves the average throughput of DAPO by 4.96\% compared to the baseline. The improvement gap becomes more pronounced as training progresses. 
% The core mechanism of \gapo, adaptively identifying an HDI and using its median for advantage estimation, adds negligible computational overhead on the GPU.

\subsection{Hyperparameter Study} 

% \begin{figure}[h]
% \centering
% \includegraphics[width=\linewidth]{figs/W_B_hyper-grpo.pdf}
% \caption{Performance curves on the test set for GRPO with $\tau$ values of 0.1, 0.5 (default), and 0.9, as logged by W\&B. }
% \label{fig:hyper_curve_per_grpo}
% \end{figure}

\begin{figure}[h]
\centering
\includegraphics[width=\linewidth]{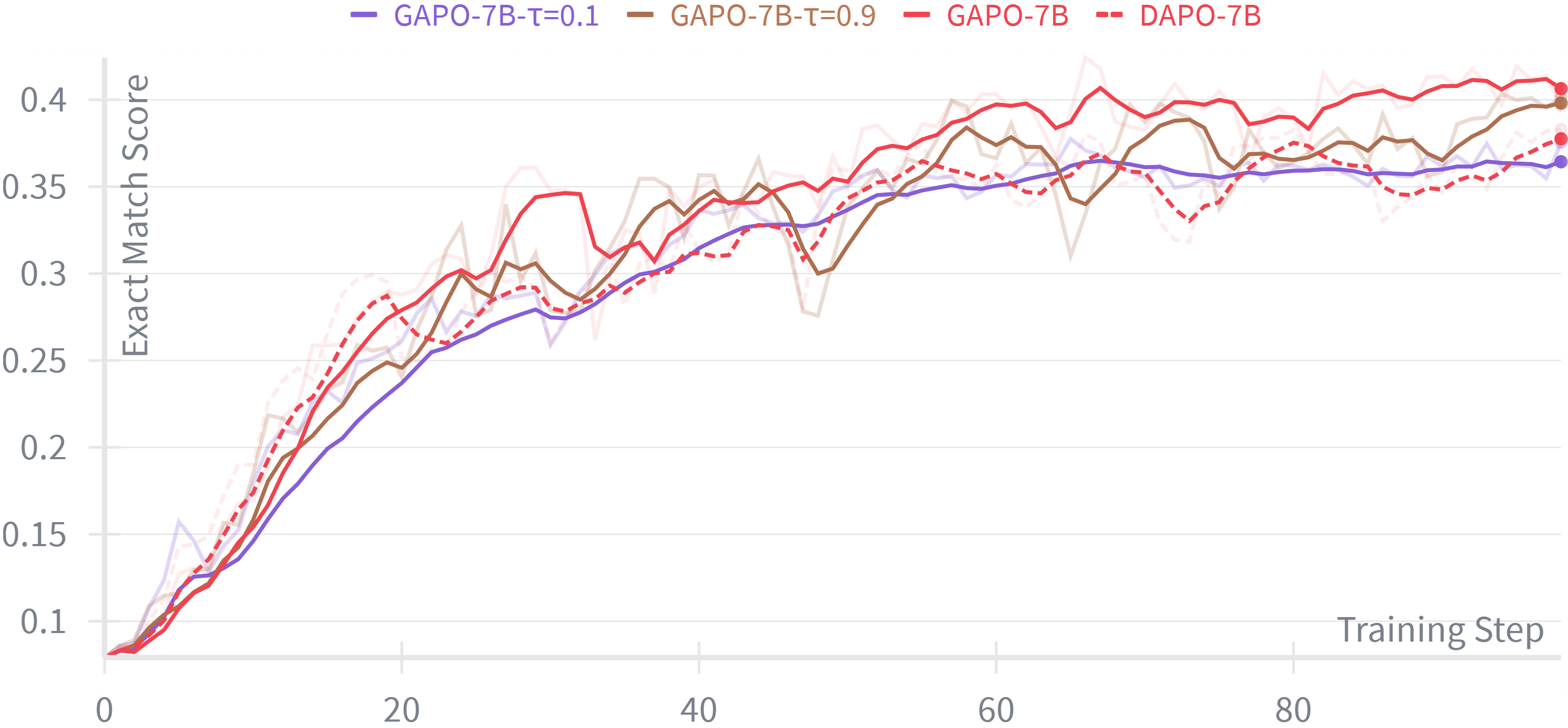}
\caption{Performance curves on the test set for DAPO with Qwen2.5-Coder-7B using $\tau$ values of 0.1, 0.5 (default), and 0.9, as logged by W\&B.  }
\label{fig:hyper_curve_per_dapo}
\end{figure}

% \begin{figure}[h]
% \centering
% \includegraphics[width=\linewidth]{figs/W_B_hyper-grpo-actor.pdf}
% \caption{Clip fraction curves on the training set for GRPO with $\tau$ values of 0.1, 0.5 (default), and 0.9, as logged by W\&B. }
% \label{fig:hyper_curve_clip_grpo}
% \end{figure}

\begin{figure}[h]
\centering
\includegraphics[width=\linewidth]{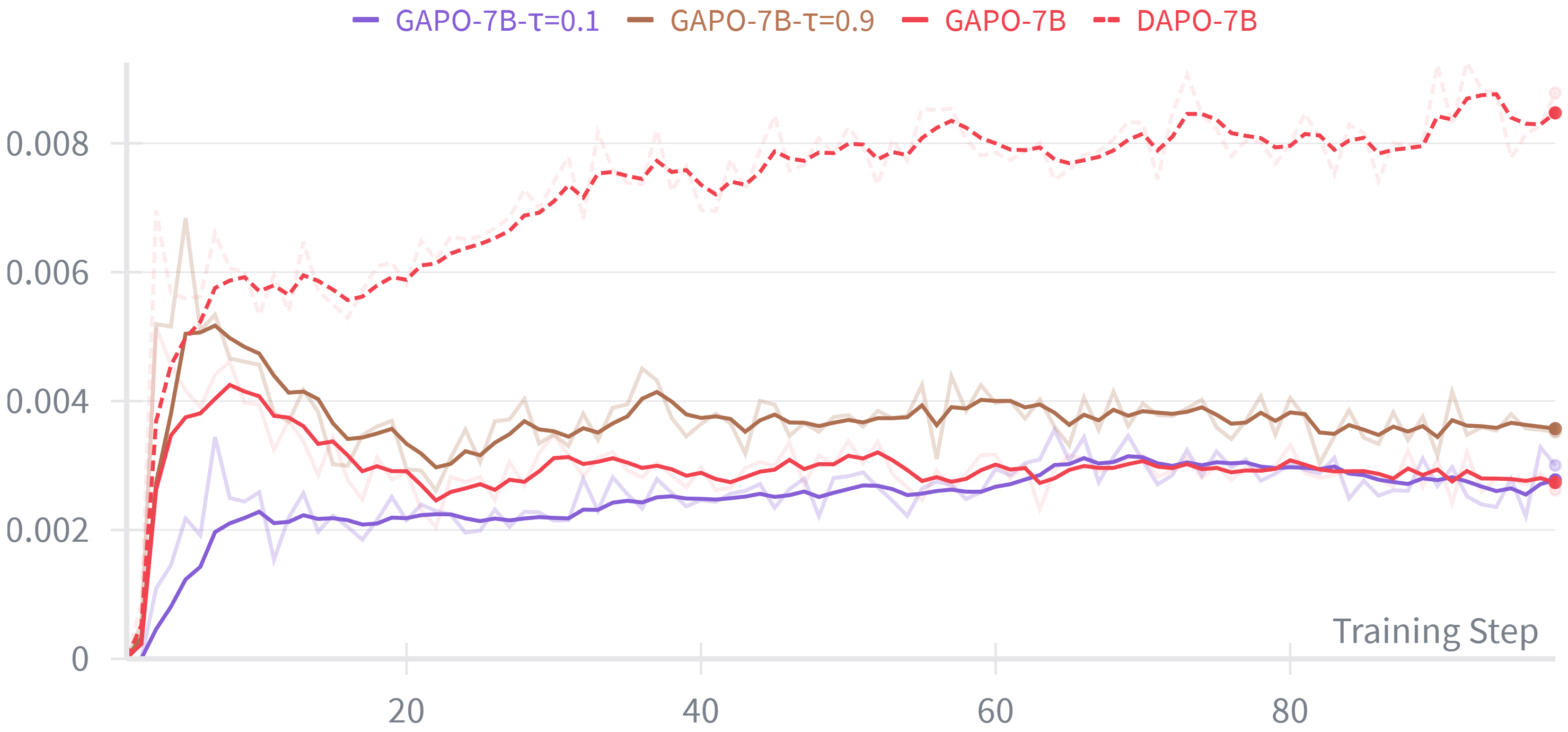}
\caption{Clip fraction curves on the training set for DAPO with Qwen2.5-Coder-7B using $\tau$ values of 0.1, 0.5 (default), and 0.9, as logged by W\&B.  }
\label{fig:hyper_curve_clip_dapo}
\end{figure}

The only hyperparameter of our \gapo method is $\tau$, which defines the percentage range of the dense region. We study its effects on Qwen2.5-Coder-7B, with the results shown in \cref{fig:curve_per_dapo} and \cref{fig:curve_clip_dapo}. Across these figures, we observe that the default value of $\tau = 0.5$ achieves the best overall performance. While $\tau = 0.9$ produces suboptimal learning curves, it exhibits higher instability, as indicated by larger \texttt{pg\_clipfrac} values in DAPO frameworks. Conversely, $\tau = 0.1$ yields lower accuracy but demonstrates the best stability, evidenced by smoother learning curves and smaller \texttt{pg\_clipfrac} values. Therefore, $\tau$ can be tuned within the range 0.1–0.5 to balance accuracy and stability according to different requirements.

\subsection{Ablation Study} 

\begin{table}[h]
\caption{Exact match accuracy of \gapo's variants on the test set using Qwen2.5-Coder (3B, 7B, 14B). }
\centering
\resizebox{\linewidth}{!}{
\begin{tabular}{l|ccc}
\toprule
Size & 3B & 7B & 14B \\
\midrule
GRPO & 39.69 & 40.05 & 42.64 \\
\gapo (G) (median, div) & 42.70 & 44.40 & 46.25 \\
\midrule
\gapo (G) (median, std) & 39.42 & 40.23 & 44.43 \\
$\quad \Delta$ to GRPO & \textcolor{red_}{\textbf{-0.27}} & \textcolor{green_}{\textbf{+0.18}} & \textcolor{green_}{\textbf{+1.79}} \\
$\quad \Delta$ to \gapo (G) & \textcolor{red_}{\textbf{-3.28}} & \textcolor{red_}{\textbf{-4.17}} & \textcolor{red_}{\textbf{-1.82}} \\
\midrule
\gapo (G) (mean, div) & 41.72 & 42.54 & 45.10 \\
$\quad \Delta$ to GRPO & \textcolor{green_}{\textbf{+2.03}} & \textcolor{green_}{\textbf{+2.49}} & \textcolor{green_}{\textbf{+2.46}} \\
$\quad \Delta$ to \gapo (G) & \textcolor{red_}{\textbf{-0.98}} & \textcolor{red_}{\textbf{-1.86}} & \textcolor{red_}{\textbf{-1.15}} \\
\midrule
DAPO & 38.74 & 41.64 & --- \\
\gapo (D) (median, div) & 39.98 & 43.96 & --- \\
\midrule
\gapo (D) (median, std) & 37.32 & 40.65 & --- \\
$\quad \Delta$ to DAPO & \textcolor{red_}{\textbf{-1.62}} & \textcolor{red_}{\textbf{-0.99}} & --- \\
$\quad \Delta$ to \gapo (D) & \textcolor{red_}{\textbf{-2.86}} & \textcolor{red_}{\textbf{-3.31}} & --- \\
\midrule
\gapo (D) (mean, div) & 38.96 & 41.11 & --- \\
$\quad \Delta$ to DAPO & \textcolor{green_}{\textbf{+0.22}} & \textcolor{red_}{\textbf{-0.53}} & --- \\
$\quad \Delta$ to \gapo (D) & \textcolor{red_}{\textbf{-1.02}} & \textcolor{red_}{\textbf{-2.85}} & --- \\
\bottomrule
\end{tabular}}
\label{tab:abl}
\end{table}

We have demonstrated the superiority of \gapo over QAE, which uses a non-adaptive quantile similar to our median; therefore, we do not perform an ablation of the adaptive HDI. 
In addition to using the median within the adaptive dense region as our adaptive $Q$, denoted \gapo (median, div), which modifies both the numerator and denominator in the original advantage computation of GRPO (\cref{eq:adv_grpo}), we consider two variants: 
\begin{enumerate}
    \item \gapo (median, std): replaces only the numerator with the median while keeping the denominator unchanged (\ie, still based on the standard deviation);
    \item \gapo (mean, div): uses the mean instead of the median within the adaptive dense region to compute $Q$.
\end{enumerate}

As shown in \cref{tab:abl}, both variants perform worse than the default \gapo, and in most cases even underperform the original GRPO/DAPO, particularly \gapo (median, std). In \gapo (median, std), only the numerator of the advantage computation differs from that of GRPO/DAPO, which introduces a shift in the advantage values. This shift cannot consistently eliminate the negative impact of outliers in the real-world code editing scenarios, like GMPO \cite{zhao2025geometric}; instead, it introduces noise and ultimately degrades performance. Larger models (\eg, 14B) appear more robust to this advantage shift. 

In contrast, replacing the median with the mean in both the numerator and denominator, \ie, \gapo (mean, div), results in less performance degradation than \gapo (median, std). This is because \gapo (mean, div) leverages statistics from the dense and highest-SNR sub-region to represent the entire (potentially outlier-contaminated) action distribution, reducing the influence of outliers. The fact that \gapo (median, div) outperforms \gapo (mean, div) demonstrates that the median better suppresses the impact of outliers than the mean, yielding a more robust advantage calculation. 

\section{Related Work}

\subsection{Stabilized Variants of GRPO}

To address the instability and limited exploration of GRPO, recent studies have proposed improvements from multiple perspectives.
For instance, \cite{wei2025redit} highlights GRPO’s gradient instability and mitigates it by dithering discrete reward signals with noise. 
Dr.~GRPO \cite{liu2025understanding} corrects training bias by jointly considering response length and question difficulty.
GSPO \cite{zheng2025group} mitigates high variance in Mixture-of-Experts models via sequence-level importance sampling. GRPO-MA \cite{wang2025grpo} reduces variance by averaging multi-answer rewards without calibration of reward baseline.
% Other works \cite{cui2025entropy, hao2025rethinking} stabilize training and alleviate entropy collapse by regulating entropy dynamics.

Another line of work enhances GRPO stability through reward baseline calibration but fails to adapt to skewed reward distributions.
GMPO \cite{zhao2025geometric} simply replaces the arithmetic mean with a geometric mean to mitigate outlier sensitivity, without adaptation to varying reward distributions.
KRPO \cite{wang2025kalman} relies on the Gaussian distribution assumption on rewards and hyperparameter tuning of the Kalman filter to estimate latent reward baselines and uncertainty, failing to address signal bias under skewed distributions. 
Moreover, QAE \cite{wu2025quantile} simply introduces a group-wise $K$-quantile baseline for entropy-safe reasoning, but it cannot identify noise-free regions or perform adaptive denoising. 

Despite these efforts, stability in LLM RL for code tasks remains underexplored. Besides, existing methods rely on discrete validation rewards from mathematical reasoning, unlike the continuous rewards in the code-editing tasks studied here. 

\subsection{LLM for Code Edit}

Code editing \cite{li2023codeeditor, guo2024codeeditorbench, nam2025prompting} is more challenging than basic code refinement, as it depends on the complex context and cursor condition.
To address this, Self-Edit \cite{zhang2023self} fine-tunes a fault-aware neural editor in a generate-and-edit manner to improve code quality and accuracy.
EDITLORD \cite{li2025editlord} avoids direct prompting or fine-tuning by using one LLM to extract transformation rules and a second to apply them to code pairs.
IterPref \cite{wu2025iterpref} leverages offline preference learning for iterative debugging, enabling context-aware code editing based on user feedback. 
LEDEX \cite{jiang2024ledex} automatically collects refinement trajectories and enhances LLMs’ self-debugging ability via supervised fine-tuning and RL.  
Finally, RLEF \cite{gehring2024rlef} performs code editing with real-time execution feedback, optimizing refinement through end-to-end RL.

\section{Conclusion}

We propose \gapo, a robust RL method that replaces the global mean with an adaptive group-wise (Q) to handle skewed, outlier-prone distributions in real-world code editing. Tested on 10 languages and nine LLMs (3B–14B), \gapo consistently outperforms alternatives with minimal overhead, improving generalization on left-skewed tasks and specialization on right-skewed ones, providing a plug-and-play solution for stable RL-based code LLM post-training.

\section{Limitations}

We focus on adaptively identifying the highest SNR region to mitigate noise when estimating the true reward distribution. However, identifying a single continuous high-SNR region may overlook multi-peak reward distributions, leading to inaccurate reward estimates. Additionally, the noise distribution is currently unknown and unpredictable, but it presents an avenue for future exploration.

\bibliography{custom}

% \newpage
\appendix

\section{Additional Experiments}

\subsection{Additional Settings}

We follow most of the default settings for GRPO and DAPO in \texttt{verl} with adaptive modifications for our code edit tasks, such as an input prompt length of 4096, an output length of 1024, a rollout batch size of 512, a training batch size of 32, 8 rollouts per iteration, and a total of 10 epochs. We use the default best hyperparameter settings for baseline methods. 
For zeta \cite{zedindustries_zeta}, we evaluate on its ``dpo'' subset, since the ``eval'' subset contains only 33 examples.

\subsection{Reward Distribution Analysis}

\begin{figure}[h]
\centering
\includegraphics[width=\linewidth]{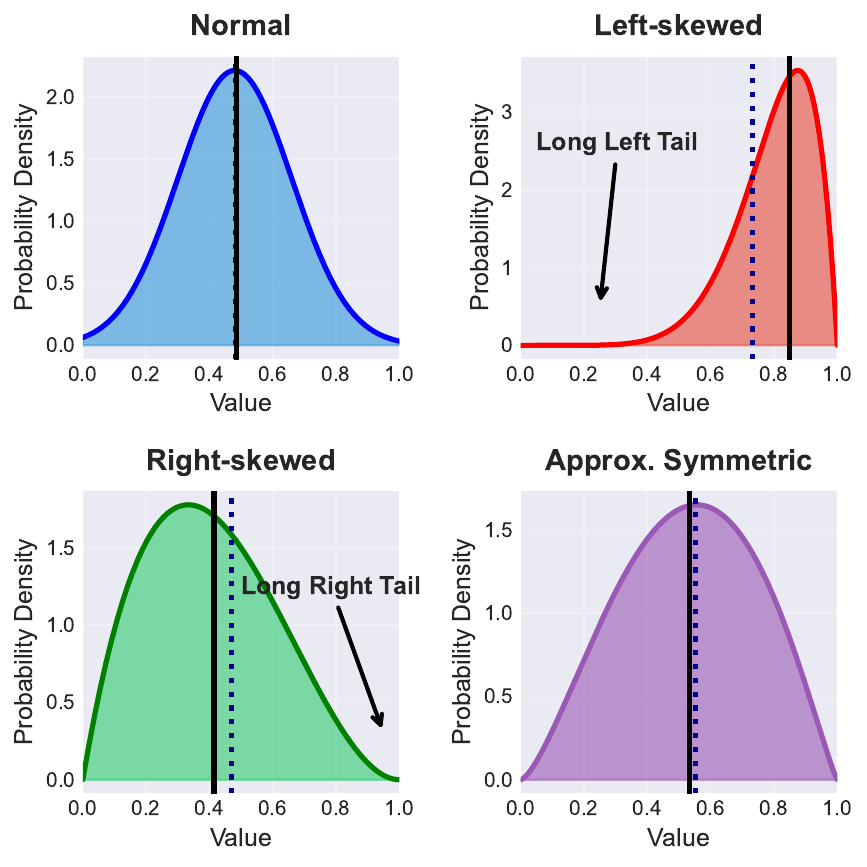}
\caption{Examples of four types of reward distribution of rollouts before training using Qwen2.5-Coder-14B.}
\label{fig:distribution}
\end{figure}

Furthermore, we visualize representative examples sampled from input prompts that exhibit four types of reward distributions in \cref{fig:distribution}, providing a closer look at the skewed cases. Mean–median gaps are evident in left- and right-skewed distributions, where outliers mainly occur. By further analyzing the relationship between the mean–median difference and the sign of the advantages computed in \cref{eq:adv_grpo} and \cref{eq:adv_our}, we find that our \gapo method generates more negative rollouts than the original GRPO/DAPO in left-skewed distributions. This occurs because $Q$ is larger than the mean in such cases, causing most rewards to fall below $Q$ and thus produce negative advantages. This behavior is reasonable, as left-skewed distributions typically correspond to \textit{relatively easy problems} with generally higher rewards, where more negative samples promote better generalization during training \cite{mu2025dissecting, zhu2025surprising}. Conversely, for right-skewed distributions (\textit{relatively hard problems}), \gapo encourages more specialized learning, leading to improved accuracy on these challenging cases. These trends align with our post-training goals. 

\end{document}